\documentclass[]{TEAI}
\usepackage{helvet}
\usepackage{xspace}
\usepackage{enumitem}
\usepackage{pifont}

\definecolor{mydarkblue}{rgb}{0.21,0.49,0.74}
\def\eg{\textit{e.g.}}
\def\ie{\textit{i.e.}}

\newcommand*{\system}{ARM\@\xspace}
\newcommand{\Checkmark}{\ding{51}}
\newcommand{\XSolidBrush}{\ding{55}}
\newcommand{\equalmark}{\textsuperscript{*}}
\newcommand{\corrmark}{\textsuperscript{\textdagger}}
\newcommand{\keywords}[1]{\par\vspace{0.5em}\noindent{\small\textbf{Keywords:} #1}\par\vspace{0.5em}}

\theoremstyle{plain}

\theoremstyle{definition}

\theoremstyle{remark}

\title{\system: An AutoRegressive Large Multimodal Model with Unified Discrete Representations}

\renewcommand{\authorlist}{%
{\authorfont\bfseries Junke Wang\equalmark},
{\authorfont\bfseries Xiao Wang\equalmark},
{\authorfont\bfseries Jiacheng Pan\equalmark},
{\authorfont\bfseries Xuefeng Hu\equalmark} \\
{\authorfont\bfseries Feng Li},
{\authorfont\bfseries Jingxiang Sun},
{\authorfont\bfseries Chaorui Deng},
{\authorfont\bfseries Zilong Chen},
{\authorfont\bfseries Yunpeng Chen},
{\authorfont\bfseries Kaibin Tian} \\
{\authorfont\bfseries Matthew Gwilliam},
{\authorfont\bfseries Hao Chen},
{\authorfont\bfseries Danhui Guan},
{\authorfont\bfseries Kun Xu},
{\authorfont\bfseries Weilin Huang} \\
{\authorfont\bfseries Zuxuan Wu\corrmark},
{\authorfont\bfseries Haoqi Fan\corrmark},
{\authorfont\bfseries Yu-Gang Jiang\corrmark},
{\authorfont\bfseries Zhenheng Yang\corrmark}%
}
\renewcommand{\affiliationlist}{%
{\affiliationfont Institute of Trustworthy Embodied AI, Fudan University\quad ByteDance TikTok\quad ByteDance Seed}%
}
\renewcommand{\contributionlist}{%
{\contributionfont \equalmark Equal contribution. \corrmark Correspondence.}%
}

\abstract{\leavevmode This paper introduces \textbf{\system}, a discrete representation-based \textbf{A}uto\textbf{R}egressive \textbf{M}odel that unifies image understanding, generation, and editing within a next-token prediction framework. \system is built on three efforts: first, we train a discrete semantic visual tokenizer that maps images into compact token sequences. Our tokenizer is supervised with multiple objectives that jointly promote semantic discriminability, language alignment and faithful reconstruction, thereby supporting diverse tasks in a shared latent space. With this, we train a 7B autoregressive model over large-scale text and image token sequences, seamlessly developing vision-language perception and generation capabilities. Finally, to further improve preference-aligned behavior for text-to-image generation and instruction-guided editing, \system applies reinforcement learning (RL) to optimize task-level objectives such as visual quality, instruction adherence, and edit consistency. Surprisingly, the results show that RL not only substantially improves performance on the target tasks (\eg, raising WISE overall \textbf{from 0.50 to 0.56}, GEdit-Bench-EN G\_O \textbf{from 5.75 to 6.68}), but also induces cross-task synergy between text-to-image generation and editing. Collectively, these findings highlight autoregressive modeling, when paired with strong representations and preference optimization, as a scalable foundation for multimodal intelligence. Project page: \href{https://github.com/wdrink/ARM}{\texttt{https://github.com/wdrink/ARM}}.
}

\begin{document}
\maketitle

\begin{figure}[!t]
\centering
\includegraphics[width=\textwidth]{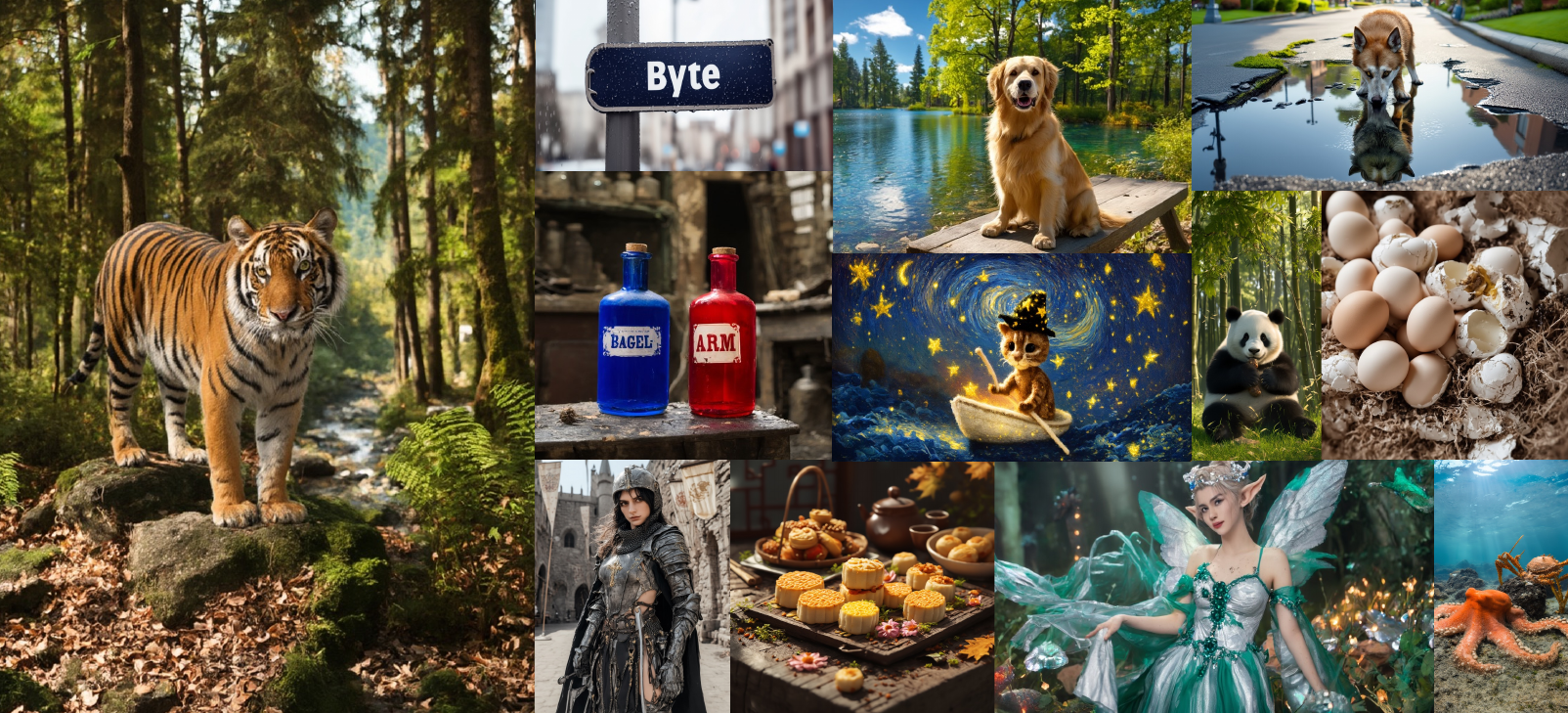}
\caption{High-resolution images of various aspect ratios generated by \system.}
\label{teaser}
\end{figure}

\section{Introduction}
\label{sec:intro}
Large multimodal models (LMMs)~\cite{alayrac2022flamingo,li2023blip,liu2023visual} have matured into a scalable paradigm for integrating visual perception with language modeling~\cite{brown2020language}, achieving consistent improvements on vision-language benchmarks and showcasing increasingly general cross-modal reasoning and instruction-following capabilities~\cite{li2024llava,bai2025qwen2,guo2025seed1}. Building on this momentum, recent work has explored extending LMMs beyond understanding toward end-to-end frameworks that unify multimodal understanding and generation~\cite{team2024chameleon,wu2024vila,xie2024show,ma2025unitok,wu2026liquid}. Representative routines include hybrid architectures that couple token prediction with denoising~\cite{zhou2024transfusion,deng2025emerging}, modular designs that pair an LMM with a separate diffusion image generator~\cite{wu2025qwen}, and fully autoregressive designs that predict text and visual tokens in a consistent manner~\cite{wang2024emu3}.

Despite the progress achieved, most existing methods~\cite{zhou2024transfusion,wu2025janus,deng2025emerging} still rely on separate visual encoders for multimodal understanding and generation to accommodate a long-standing mismatch in the visual representations favored by the two tasks. While technically feasible, this design leaves the system structurally fragmented, since the model must devote additional modeling capacity to bridging two distinct visual latent spaces~\cite{deng2025emerging}. Moreover, redundant representations of the same visual input must be carried in the context and jointly consumed by the model in cross-modal reasoning and interleaved generation, incurring substantial overhead during inference~\cite{deng2025emerging,liao2025mogao}. Although a few recent efforts attempt to unify understanding and generation with generation-oriented visual tokens~\cite{team2024chameleon,wang2024emu3}, they significantly compromise understanding performance to prioritize synthesis fidelity. 

To address these issues, we propose \system, a large multimodal model with unified discrete representations. The core of \system is a discrete visual tokenizer trained with complementary supervision signals, encouraging it to preserve both text-aligned semantics for recognition and appearance details for high-fidelity synthesis and editing. Building on this tokenizer, we train a 7B autoregressive model over large-scale interleaved text and visual token sequences, developing unified capabilities for vision-language understanding, image generation, and instruction-guided editing. Finally, to better align \system with user preferences, we further improve it with Group Relative Policy Optimization~\cite{guo2025deepseek}, using powerful multimodal models, \ie, GPT~\cite{achiam2023gpt}, as reward models. Benefiting from the discrete visual token design, we surprisingly find that this stage induces cross-task synergy: optimizing either text-to-image generation or editing consistently benefits the other, and joint training yields further gains. More importantly, multimodal understanding performance remains stable, suggesting that the alignment of generative preferences for visual token prediction does not degrade the inherent understanding capacity of the model.

With the above efforts, \system delivers state-of-the-art or competitive performance across multimodal understanding, generation, and editing. For example, we achieve 40.2 and 87.3 on MMMU~\cite{yue2024mmmu} and POPE~\cite{li2023evaluating}, respectively, substantially outperforming prior methods that rely on discrete visual representations. For image generation, \system reaches 0.86 and 0.56 on GenEval~\cite{ghosh2023geneval} and WISE~\cite{niu2025wise}, attaining leading-level results relative to diffusion baselines~\cite{bfl_flux_2024}. In terms of image editing, \system achieves strong results on GEdit-Bench-EN~\cite{liu2025step1x}, with a G\_O score of 6.68 on the full set. These results indicate the potential of autoregressive modeling in multimodal artificial intelligence, paired with strong representations and effective training.

\section{Related Work}
\label{sec:related}
\noindent \textbf{Unified Visual Tokenizer.} Two primary categories of visual tokenizers have emerged for converting images into 1D sequences. 
Semantic visual encoders like CLIP~\cite{radford2021learning,bolya2025perception} and SigLIP~\cite{zhai2023sigmoid,tschannen2025siglip} preserve high-level representations, enhancing visual understanding in MLLMs but failing to capture fine-grained details needed for precise image generation and editing.
Fine-grained visual encoders such as VQVAEs~\cite{esser2021taming,wang2024omnitokenizer} and VAEs~\cite{rombach2022high} excel at visual generation through reconstruction-based training but exhibit weaker semantic alignment for understanding tasks.
Recent unified approaches~\cite{wu2024vila,ma2025unitok,han2025vision,lu2025atoken} address these limitations by supporting both tasks. VILA-U~\cite{wu2024vila}, UniTok~\cite{ma2025unitok}, and AToken~\cite{lu2025atoken} jointly optimize image–text alignment and reconstruction, while TAR~\cite{han2025vision} reconstructs the SigLIP~\cite{zhai2023sigmoid} feature space via discrete quantization.

\noindent \textbf{Unified Vision Language Model.} Following MLLM visual understanding success~\cite{li2024llava,li2023blip}, studies~\cite{xie2024towards} increasingly explore unified MLLMs for both understanding and generation.
Early approaches like Next-GPT~\cite{wu2024next}, SEED-X~\cite{ge2024seed}, and EMU2~\cite{sun2024generative} use semantic encoders to tokenize images, with separate diffusion models generating final images from MLLM outputs. 
While performing well on understanding and generation, they struggle with editing due to lost fine-grained details.
Another line~\cite{wang2024emu3,team2024chameleon,liu2024world} adopts VQ-GAN–style architectures~\cite{esser2021taming}, using encoders for understanding and decoders for generation. However, lacking semantically aligned features limits their understanding performance.
Recent unified-tokenizer approaches~\cite{wu2024vila,ma2025unitok,han2025vision} learn shared representations for both tasks but struggle to achieve optimal performance.
Alternatively, works like Janus-Pro~\cite{chen2025janus}, Bagel~\cite{deng2025emerging}, Mogao~\cite{liao2025mogao} decouple visual encoding using separate encoders. 
Though achieving strong results, this substantially increases computational cost for editing by requiring two distinct visual embeddings.

\noindent \textbf{Visual Generation.} Visual generation produces high-fidelity images from textual or multimodal input through three main approaches: (1) Autoregressive (AR) models~\cite{esser2021taming,sun2024autoregressive,wang2025simplear,wang2025omnigen} advanced generation by mapping images to discrete tokens via refined VQ-VAE~\cite{yu2023magvit,guo2025dera} architecture. MLLMs~\cite{yu2022scaling,team2024chameleon,sun2024autoregressive,wang2025simplear} and unified MLLMs~\cite{wu2024vila,ma2025unitok,lu2025atoken} leverage such discrete tokens for autoregressive generation with strong results. (2) Masked prediction models~\cite{chang2022maskgit,yu2023magvit} generate VQ tokens in parallel. Modern MLLMs~\cite{chang2023muse,xie2024show,tian2025unigen,tian2026unigen} incorporating these methods achieve superior generation performance among discrete unified MLLMs. (3) Diffusion models~\cite{ho2020denoising,song2020denoising,peebles2023scalable} surpass VQ-based approaches in fidelity and diversity. 
Operating in continuous VAE latent spaces~\cite{kingma2013auto} further elevated quality~\cite{rombach2022high,esser2024scaling,brooks2024video,xie2024sana,bfl_flux_2024}. 
Recent MLLMs~\cite{wu2024next,ge2024seed,chen2025blip3,wang2025growing} integrate latent diffusion models to decode visual outputs, while emerging approaches~\cite{liao2025mogao,deng2025emerging,xie2025show} employ LLMs directly within diffusion processes.

\section{Methods}
\label{sec:methods}
\system adopts a single autoregressive transformer backbone, where multimodal inputs are tokenized into one-dimensional discrete sequences via their respective text and visual tokenizers. These interleaved sequences are then modeled with next-token prediction. Finally, modality-specific detokenizers map the predicted discrete tokens back to natural languages or pixels.

Next, we introduce the key components of \system and its training pipeline. First, Sec.~\ref{subsec:unified_tokenizer} presents the unified discrete visual tokenizer that bridges images and discrete sequence modeling. Sec.~\ref{sec:stage3} then describes large-scale autoregressive training over interleaved text and visual tokens. Finally, Sec.~\ref{sec:stage4} outlines preference-based reinforcement learning, which further aligns the prediction of visual tokens with human feedback.

\subsection{Unified Discrete Visual Tokenization}
\label{subsec:unified_tokenizer}
The foundation to enable visual understanding, generation, and editing through a single autoregressive backbone is a visual tokenizer that retains both high-level semantics and fine-grained visual details. Our unified tokenizer is built on a pretrained SigLIP2 encoder~\cite{tschannen2025siglip}, which provides semantically strong visual features for discretization. The SigLIP2 backbone remains frozen during training to preserve its representative capability and stabilize optimization. On top of the encoder outputs, a projection module implemented as stacked attention blocks maps the high-dimensional embeddings into a compact latent subspace.

\begin{figure}[t]
\centering
\includegraphics[width=0.8\linewidth]{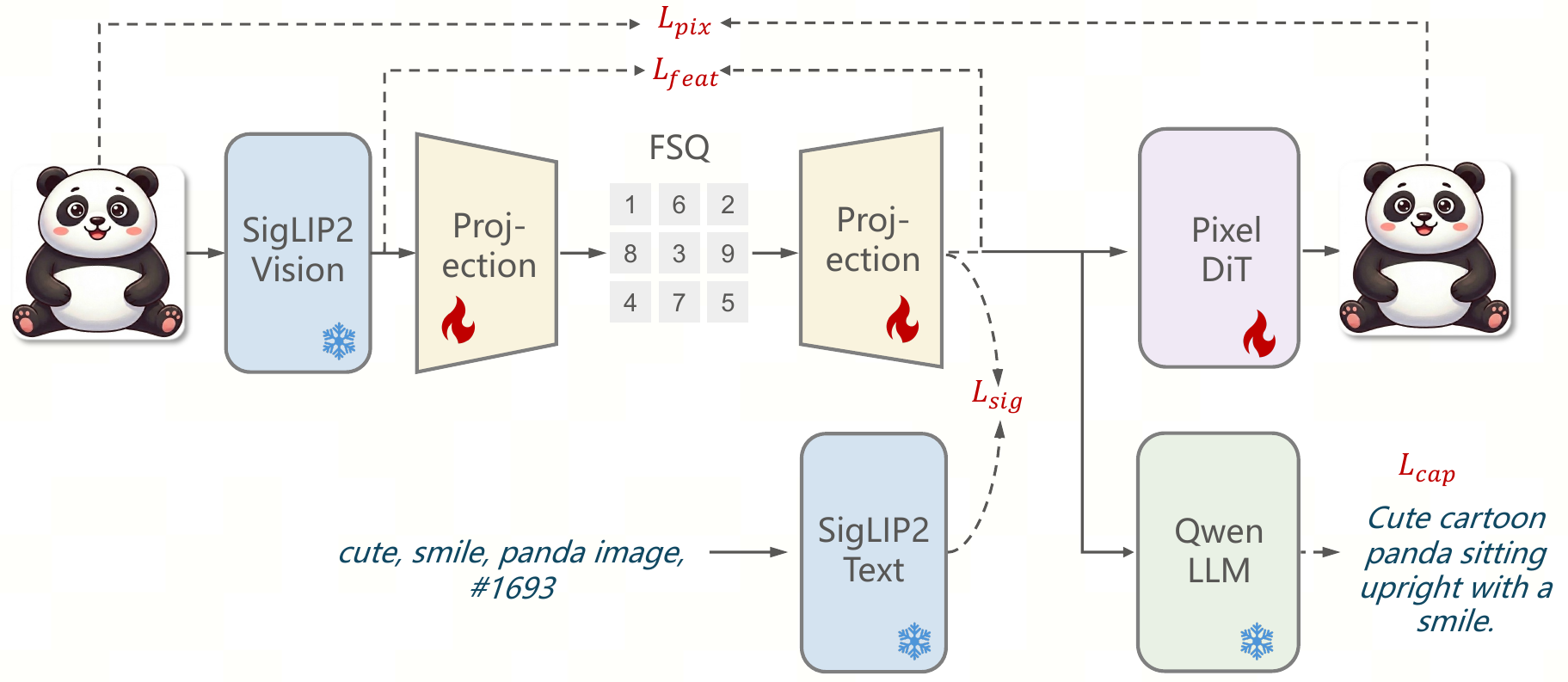}
\caption{Architecture of our unified discrete visual tokenizer.}
\label{fig:tokenizer_diagram}
\end{figure}

Discretization is performed with Finite Scalar Quantization (FSQ)~\cite{mentzer2023finite}, which offers high capacity without the need of an explicit codebook. A symmetric projection module follows quantization, mapping the quantized embeddings back to the original feature dimension. The overall tokenizer architecture is illustrated in Figure~\ref{fig:tokenizer_diagram}.

Our tokenizer is supervised with four complementary objectives that jointly promote semantic alignment and reconstruction fidelity, as described below.

\noindent \textbf{1) Caption loss}: to align the tokenizer representations with the language model~\cite{team2024qwen2}, we adopt a captioning objective, $\mathcal{L}_{\text{cap}}$, formulated as a cross-entropy over text tokens $y_i$ conditioned on the quantized visual representation $z_q$ and the preceding context $y_{<N}$: 
\begin{equation}
L_{\text{cap}} = -\sum_{i=1}^{N}\log p_{\phi}(y_i|z_q,y_{<i})
\label{loss:caption}
\end{equation}
where $\phi$ denotes a pre-trained language model~\cite{team2024qwen2}, $N$ is the length of text tokens.

\noindent \textbf{2) Pixel reconstruction loss}: to preserve the low-level information required for high-fidelity synthesis, we define a pixel reconstruction loss $\mathcal{L}_{\text{pix}}$ by training a lightweight diffusion transformer decoder $D_{\text{pix}}$~\cite{peebles2023scalable} to learn the rectified velocity field~\cite{lipman2022flow} in pixel space:
\begin{equation}
\mathcal{L}_{\text{pix}} = \mathbb{E}_{t, x_0, x_1} \left\| D_{\text{pix}}(x_t, t \mid z_q) - (x_1 - x_0) \right\|_2^2,
\label{loss:recon}
\end{equation}
where $x_1$ denotes the target image, $x_0$ is sampled Gaussian noise, and $x_t = t x_1 + (1-t) x_0$ is the linear interpolation with $t \sim \mathcal{U}[0,1]$. Optimizing directly in pixel space rather than in a VAE~\cite{pu2016variational} latent space avoids lossy compression from the VAE bottleneck, which helps preserve appearance fidelity under quantization. In addition, our diffusion decoder enjoys more stable optimization compared to GAN-style decoders~\cite{goodfellow2014generative}.

\noindent \textbf{3) Sigmoid loss}: we further introduce a sigmoid contrastive objective~\cite{zhai2023sigmoid} $\mathcal{L}_{\text{sig}}$ to align the quantized visual embedding $z_q$ with the corresponding SigLIP2 text embedding $s$~\cite{tschannen2025siglip}:
\begin{equation}
\begin{split}
\mathcal{L}_{\text{sig}} = &-\log \sigma\!\left(\tau\,\cos(z_q, s)+b\right) \\
&-\sum_{s_j\in \mathcal{B},\, s_j\neq s}\log \sigma\!\left(-\left(\tau\,\cos(z_q, s_j)+b\right)\right),
\end{split}
\label{loss:contrastive}
\end{equation}
where $\mathcal{B}$ denotes the set of text embeddings in the current batch, $\tau$ and $b$ are learnable scalars that scale and shift the logits. $\sigma(\cdot)$ is the sigmoid function, and $\cos(\cdot,\cdot)$ denotes cosine similarity.

\noindent \textbf{4) Feature distillation loss}: finally, we use a feature distillation loss $\mathcal{L}_{\text{feat}}$ to match the quantized embeddings $z_{q}$ with original SigLIP2 visual features $z$ by minimizing their cosine distance.

The final optimization objective for the unified discrete visual tokenization $L_{\text{Tok}}$ combines the above components with balancing weights:
\begin{equation}
L_{\text{Tok}} = \lambda_{cap} L_{\text{cap}} + \lambda_{pix} L_{\text{pix}} + \lambda_{sig} L_{\text{sig}} + \lambda_{feat} L_{\text{feat}},
\label{loss:overall}
\end{equation}
where $\lambda_{cap}$, $\lambda_{pix}$, $\lambda_{sig}$, and $\lambda_{feat}$ are set to 1, 5, 5, 1.

\textbf{Detokenization via Latent Diffusion Decoder}. While the lightweight pixel-space decoder $D_{\text{pix}}$ provides supervision for detail preserving, a separate high-capacity latent diffusion model~\cite{bfl_flux_2024} is used for high-quality detokenization, conditioned on the learned quantized embeddings.

Concretely, we start from a pretrained latent DiT model $D_{\text{latent}}$ and replace its text conditioning with $z_q$ produced by our tokenizer. $D_{\text{latent}}$ is trained to transport Gaussian noise $z_0 \sim \pi_0$ to target image latents $z_1 \sim \pi_1$, using a rectified-flow objective~\cite{liu2022flow}:
\begin{equation}
\mathcal{L}_{\text{Detok}} = \mathbb{E}_{t, z_0, z_1} \left\| d_{\text{latent}}(z_t, t \mid z_q) - (z_1 - z_0) \right\|_2^2,
\label{loss:diff_mse}
\end{equation}
where $z_1$ is obtained by encoding the target image with a pretrained VAE~\cite{bfl_flux_2024}.

\subsection{Autoregressive Large Multimodal Model}
\label{sec:stage3} 
The discrete visual tokenizer detailed in Section~\ref{subsec:unified_tokenizer} allows us to represent all visual inputs and outputs as discrete tokens. With this, we pack diverse data of different modalities and tasks (\eg, image-to-text, text-to-image, text-only, and interleaved image-text) into flattened multimodal token sequences, and model their dependencies via a standard next-token prediction objective:
\begin{equation}
L_{\text{\system}} = -\sum_{j=1}^{M}\log p_{\theta}(y_j|y_{<j}),
\label{loss:vlm}
\end{equation}
where $M$ denotes the total sequence length, and our autoregressive LMM is parameterized by $\theta$.

\subsection{Preference Alignment for Visual Token Prediction}
\label{sec:stage4}
Large-scale multimodal next-token prediction training provides a strong foundation for \system to perform unified understanding, generation, and editing. Building on this foundation, we further employ Group Relative Policy Optimization (GRPO)~\cite{shao2024deepseekmath} to directly align the model with preference feedback. Note that during this stage, optimization is applied only to the visual token prediction, targeting generation and editing downstream tasks.

We initialize the policy $\pi_\theta$ from the previous stage and set the reference policy $\pi_{\text{ref}}$ as a frozen copy of the same checkpoint throughout GRPO. Given a prompt $x$, $\pi_\theta$ samples a group of $K$ visual token sequences $\{y^{1}, \dots, y^{K}\}$, which are then detokenized into images using the latent diffusion decoder. After this, we score different images with a reward model to yield the corresponding rewards $\mathbf{r}=\{r_1,\dots,r_K\}$. Advantages are computed by normalizing rewards within the group, $A_k = (r_k - \text{mean}(\mathbf{r})) / \text{std}(\mathbf{r})$. $\pi_\theta$ is updated with the following objective:
\begin{equation}
\begin{aligned}
\mathcal{L}_{\text{GRPO}} = \frac{1}{K} \sum_{k=1}^K \bigg\{ &\min \Big[ \rho_k A_k, \text{clip}(\rho_k, 1-\epsilon, 1+\epsilon)\, A_k \Big] \\
&- \beta \,\mathcal{D}_{\text{KL}}\!\left[\pi_\theta(y^{k}\!\mid x)\,\|\,\pi_{\text{ref}}(y^{k}\!\mid x)\right] \bigg\},
\end{aligned}
\label{eq:grpo}
\end{equation}
where $\rho_k = \frac{\pi_\theta(y^{k}\!\mid x)}{\pi_{\text{old}}(y^{k}\!\mid x)}$ is the probability ratio and $\beta$ controls the strength of the KL regularization.

\section{Experiments}
\subsection{Experimental Setup}

\noindent \textbf{Implementation Details.} Our tokenizer comprises a frozen SigLIP2-SO400M-512 encoder~\cite{tschannen2025siglip}, an FSQ quantizer~\cite{mentzer2023finite}, and two lightweight projection modules. The quantizer uses $L_i=2$ for $1 \le i \le 16$, corresponding to a 65K codebook. The projection modules are implemented as 6 transformer blocks. The pixel diffusion model in Eq.\ref{loss:recon} follows a DiT architecture~\cite{peebles2023scalable} with 24 transformer blocks, while the language model in Eq.~\ref{loss:caption} is a frozen 0.5B Qwen2.5~\cite{team2024qwen2}. The latent diffusion model in Eq.\ref{loss:diff_mse} is initialized from FLUX.1 [dev]~\cite{bfl_flux_2024}.

We train the tokenizer on 2.2B internal image-text pairs using AdamW~\cite{loshchilov2017decoupled} with learning rate $3 \times 10^{-4}$, $\beta_1=0.9$, and $\beta_2=0.95$. The global batch size is 32,768. 

For the unified large multimodal model, we initialize it from Qwen2.5-7B~\cite{team2024qwen2}, and append an additional linear layer for visual tokens prediction. We support dynamic resolution image generation and editing by inserting the shape tokens into the text prompt, which explicitly specify the target height and width in the discrete token grid. 

The complete training proceeds in four stages: 1) \textbf{Pre-training:} The LLM backbone is trained on 2.5T multimodal tokens, processing images at native resolution within defined dimension constraints. 2) \textbf{Continual Training:} We train the model on 2.5T tokens by incorporating higher visual resolutions and increasing the sampling ratio of interleaved data to improve reasoning capabilities. 3) \textbf{Supervised Fine-tuning:} We utilize 0.2B tokens from high-quality instruction-following datasets to strengthen the understanding capabilities in response to diverse user prompts, while maintaining the image generation and editing performance. 4) \textbf{Reinforcement Learning}: We implement training with VeRL~\cite{sheng2025hybridflow}. For text-to-image generation, we employ GPT-o3~\cite{openai_o3_o4mini_system_card_2025} as the reward model to inspect the object appearance, attributes, and spatial relationships of the generated images. For editing, we utilize GPT-4.1~\cite{openai_gpt41_2025} as the reward model, which evaluates the edited images from instruction following, preservation of non-target regions, and overall visual quality.

\begin{itemize}[leftmargin=*]
\item \textbf{Text-only Data:} as strong language modeling underpins cross-modal reasoning and instruction following, we include a curated collection of high-quality text-only data spanning general-purpose text, mathematics, and code, together with other reasoning-intensive domains.
\item \textbf{Image-to-Text Data:} we collect large-scale image-text pairs for visual understanding, primarily from web captions and alt-text. In addition to standard vision language model (VLM) datasets~\cite{li2024llava}, we incorporate OCR-rich documents, charts, and grounding-style annotations, to improve text reading and spatial understanding.
\item \textbf{Text-to-Image Data:} for text-to-image training, we use a curated collection of high-quality image-text pairs spanning diverse prompt styles. A small amount of synthetic pairs generated by existing T2I models~\cite{bfl_flux_2024,hurst2024gpt,gemini2_flash_native_image_2025,seedream2025seedream} are included to further expand stylistic coverage while maintaining visual fidelity.
\item \textbf{Interleaved Multimodal Data.} To support long-context multimodal modeling and interleaved generation with visual references, we include interleaved image-text data from video sequences~\cite{wang2025koala,han2024mvimgnet2} and web documents~\cite{commoncrawl_2007,li2024omnicorpus}. We also incorporate public image-editing datasets~\cite{bai2024humanedit,ge2024seed,hui2024hq,wei2024omniedit,zhao2024ultraedit,xiao2025omnigen} to further strengthen the capability on specific tasks.
\item \textbf{Text-to-Image RL prompts:} for text-to-image RL, we use a curated collection of mixed-format image generation prompts, featuring both short compositional prompts synthesized from ImageNet~\cite{deng2009imagenet} class names and long, dense, detailed prompts from Share-GPT-4o~\cite{chen2025sharegpt};
\item \textbf{Image Editing RL Data:} for image editing RL, we use a curated collection of image editing prompts from HQ-Editing-6000~\cite{hui2024hq} and Share-GPT-4o~\cite{chen2025sharegpt}.
\end{itemize}

For fair comparison across generation benchmarks, we run text-to-image inference at $1024\times1024$ resolution on GenEval~\cite{ghosh2023geneval}, DPG~\cite{hu2024ella}, and WISE~\cite{niu2025wise}. For GEdit~\cite{liu2025step1x}, we keep the original image resolution to match the benchmark protocol.
For text-to-image generation, we use classifier-free guidance (CFG) of 1.5 in the autoregressive model. For image editing, we apply two-branch guidance that separately conditions on the text instruction and the reference image, using guidance scales of 1.5 (text) and 1.25 (image), respectively. The diffusion decoder performs detokenization with 28 sampling steps and a CFG scale of 1.5.

\begin{table}[!ht]
\centering
\begin{minipage}[t]{0.49\linewidth}
\centering
\scriptsize
\caption{Training configuration for the large multimodal model. PT denotes pretraining, CT denotes continued training, and SFT denotes supervised fine-tuning.}
\label{tab:lmm_train_config}
\setlength{\tabcolsep}{2pt}
\renewcommand{\arraystretch}{1.26}
\resizebox{\linewidth}{!}{%
\begin{tabular}{@{}lccc@{}}
\toprule
 & PT & CT & SFT \\
\midrule
\multicolumn{4}{l}{\textbf{Hyperparameters}} \\
learning rate & $2 \times 10^{-4}$ & $2 \times 10^{-4}$ & $5 \times 10^{-5}$ \\
scheduler & Constant & Constant & Constant \\
weight decay & 0 & 0 & 0 \\
gradient norm clip & 1.0 & 1.0 & 1.0 \\
optimizer & \multicolumn{3}{c}{AdamW ($\beta_1=0.9$, $\beta_2=0.95$)} \\
warmup steps & 2K & 2K & N/A \\
training steps & 100K & 100K & 40K \\
\# of GPUs & 200 & 160 & 80 \\
max seq. length & 80K & 80K & 80K \\
T2I resolution & (256, 512) & (512, 1024) & (512, 1024) \\
Editing resolution & (256, 980) & (512, 980) & (512, 1024) \\
\midrule
\multicolumn{4}{l}{\textbf{Data sample ratio}} \\
Text-only & 0.05 & 0.05 & 0.05 \\
Image2Text & 0.10 & 0.10 & 0.05 \\
Text2Image & 0.70 & 0.55 & 0.50 \\
Interleaved gen. (video) & 0.10 & 0.15 & 0.20 \\
Interleaved gen. (web) & 0.05 & 0.15 & 0.20 \\
\bottomrule
\end{tabular}%
}
\end{minipage}\hfill
\begin{minipage}[t]{0.49\linewidth}
\centering
\scriptsize
\caption{Reinforcement Learning Parameters.}
\label{tab:lmm_rl_config}
\setlength{\tabcolsep}{2pt}
\renewcommand{\arraystretch}{1.18}
\resizebox{0.88\linewidth}{!}{%
\begin{tabular}{@{}lccc@{}}
\toprule
 & T2I RL & Editing RL & Joint RL \\
\midrule                    
\multicolumn{4}{l}{\textbf{Training Configuration}} \\
Optimizer           
& \multicolumn{3}{c}{AdamW ($\beta_1=0.9$, $\beta_2=0.95$)} \\
Learning Rate       
& 3e-5 & 5e-5 & 5e-5 \\
Scheduler           
& Constant & Constant & Constant \\
Weight Decay        
& 0.01 & 0.01 & 0.01 \\ 
Training Steps      
& 280 & 100 & 200 \\
\# of GPUs           
& 8 & 40 & 40 \\
Batch Size          
& 64 & 40 & 40 \\
\midrule
\multicolumn{4}{l}{\textbf{Sampling Configuration}} \\
Rollout        
& 16 & 16 & 16 \\
Temperature 
& 0.7 & 1.0 & 1.0 \\
Top-K 
& 1000 & 1000 & 1000 \\
Top-P 
& 1.0 & 1.0 & 1.0 \\
\midrule
\multicolumn{4}{l}{\textbf{Image Generation and Reward}} \\
Resolution          
& 512$\sim$1024 & 512$\sim$1024 & 512$\sim$1024 \\
Reward Model 
& GPT-o3 & GPT-4.1 & Mixed \\
\midrule
\multicolumn{4}{l}{\textbf{RL Configuration}} \\
KL Coef.            
& 0.01 & 0.01 & 0.01 \\
Clip Ratio          
& 0.2 & 0.2 & 0.2 \\
\bottomrule
\end{tabular}%
}
\end{minipage}
\end{table}

\subsection{Image Understanding Results}
\label{subsec:understanding}

\begin{table*}[t]
\centering
\small
\setlength{\tabcolsep}{0pt}
\renewcommand{\arraystretch}{1.1}
\caption{Multimodal understanding benchmarks. \textbf{Unified} indicates whether the model is trained for understanding only or unified understanding and generation, \textbf{\# Params} reports the size of the language model backbone.}
\label{tab:vlm_benchmark}
\begin{tabular*}{\linewidth}{@{\extracolsep{\fill}}l c c c c c c c c c c@{}}
\toprule
\textbf{Model} & \textbf{Unified} & \textbf{\#Params} & \textbf{POPE} & \textbf{MMB} & $\textbf{MME}_{\text{Perc}}$  & \textbf{MMMU} & \textbf{GQA} & \textbf{VQAv2} & \textbf{SEED} \\
\midrule
\multicolumn{10}{l}{\textit{Continuous visual representations}} \\
LLaVA-OV~\cite{li2024llava} 
& \XSolidBrush & 7B  & -    & 80.8 & 1580  & 48.8 & -    & -    & -    \\
Qwen2.5-VL~\cite{team2024qwen2} 
& \XSolidBrush & 7B  & -    & 83.5 & -     & 58.6 & -    & -    & -    \\
InternVL2.5~\cite{chen2024expanding} 
& \XSolidBrush & 8B  & -    & 84.6 & -     & 56.0 & -    & -    & -    \\
Janus-Pro~\cite{chen2025janus}      
& \Checkmark   & 7B  & 87.4 & 79.2 & 1567  & 41.0 & 62.0 & -    & 72.1 \\
BLIP-3o~\cite{chen2025blip3}       
& \Checkmark   & 8B  & -    & 83.5 & 1683  & 50.6 & -    & 83.1 & 77.5 \\
Show-o2~\cite{xie2025show}     
& \Checkmark   & 7B  & -    & 79.3 & 1621  & 48.9 & 63.1 & -    & 69.8 \\
Bagel~\cite{deng2025emerging}  
& \Checkmark   & 7B  & -    & 85.0 & 1687  & 55.3 & -    & -    & -    \\
\midrule
\multicolumn{10}{l}{\textit{Discrete visual representations}} \\
LWM~\cite{liu2024world}            & \Checkmark & 7B   & 75.2 & -    & -    & -    & 44.8 & 55.8 & -    \\
Chameleon~\cite{team2024chameleon} & \Checkmark & 34B  & -    & -    & -    & 22.4 & -    & 69.6 & -    \\
Show-o~\cite{xie2024show}          & \Checkmark & 1.3B & 80.0 & -    & 1097 & 26.7 & 58.0 & 69.4 & -    \\
Liquid~\cite{wu2026liquid}         & \Checkmark & 7B   & 83.2 & -    & 1448 & -    & 61.1 & 76.8 & -    \\
VILA-U~\cite{wu2024vila}           & \Checkmark & 7B   & 85.8 & -    & 1402 & -    & 60.8 & 79.4 & 59.0 \\
UniTok~\cite{ma2025unitok}         & \Checkmark & 7B   & 83.2 & -    & 1448 & -    & 61.1 & 76.8 & -    \\
Emu3~\cite{wang2024emu3}           & \Checkmark & 8B   & 85.2 & 58.5 & -    & 31.6 & 60.3 & 75.1 & 68.2 \\
\rowcolor[RGB]{234,242,255}
\textbf{\system} & \Checkmark                   & 7B   & 87.3 & 80.7 & 1463 & 40.2 & 59.8 & 76.1 & 73.1 \\
\bottomrule
\end{tabular*}
\end{table*}

Overall, Table~\ref{tab:vlm_benchmark} shows that \system achieves competitive understanding performance while maintaining a unified, fully autoregressive architecture.
Discrete unified models have historically lagged behind continuous unified counterparts on general understanding benchmarks, reflecting the difficulty of retaining fine-grained perceptual cues when discretizing visual signals. 
\system narrows this gap. 

\begin{table*}[!ht]
\centering
\small
\setlength{\tabcolsep}{0pt}
\renewcommand{\arraystretch}{1.1}
\caption{Comparison with state-of-the-art models on GenEval and DPG. We include diffusion models (Diff.), autoregressive models (AR), and non-autoregressive models (NAR).}
\label{tab:geneval_dpg}
\begin{tabular*}{\linewidth}{@{\extracolsep{\fill}}l c c c c c c c c@{}}
\toprule
\textbf{Methods} & \textbf{Type} &
\multicolumn{4}{c}{\textbf{GenEval}$\uparrow$} &
\multicolumn{3}{c}{\textbf{DPG}$\uparrow$} \\
\cmidrule(lr){3-6}\cmidrule(lr){7-9}
 & & \textbf{Two Obj.} & \textbf{Position} & \textbf{Color Attri.} & \textbf{Overall} &
\textbf{Global} & \textbf{Relation} & \textbf{Overall} \\
\midrule
Chameleon~\cite{team2024chameleon}   & AR    & -    & -    & -    & 0.39 & -     & -     & -     \\
PixArt-alpha~\cite{chen2023pixart}   & Diff. & 0.50 & 0.08 & 0.07 & 0.48 & 74.97 & 82.57 & 71.11 \\
Emu3~\cite{wang2024emu3} & AR    & 0.71 & 0.17 & 0.21 & 0.54 & 85.21 & 90.22 & 80.60 \\
Janus~\cite{wu2025janus}                   & AR    & 0.68 & 0.46 & 0.42 & 0.61 & 82.33 & 85.46 & 79.68 \\
SimpleAR~\cite{wang2025simplear} & AR & 0.90 & 0.28 & 0.45 & 0.63 & 87.97 & 88.33 & 81.97 \\
SD3 Medium~\cite{esser2024scaling} & Diff. &  0.74  & 0.34 & 0.36 & 0.62 & 87.90 & 80.70 & 84.08 \\
FLUX.1 [Dev]~\cite{bfl_flux_2024} & Diff. &  0.81 & 0.22 & 0.45 & 0.66 & 74.35 & 90.87 & 83.84 \\
DALL-E-3~\cite{openai_dalle3_2023}               & Diff. & 0.87 & 0.43 & 0.45 & 0.67 & 90.97 & 90.58 & 83.50 \\
Show-o~\cite{xie2024show}                  & NAR   & 0.80 & 0.31 & 0.50 & 0.68 & -     & -     & 67.48 \\
Infinity~\cite{han2025infinity}             & NAR   & 0.85 & 0.49 & 0.57 & 0.73& 93.11 & 90.76 & 83.46 \\
Lumina-Image 2.0~\cite{qin2025lumina} & Diff. &  0.87 & - & 0.62 & 0.73 & - & 94.85 & 87.20 \\
Janus-Pro-7B~\cite{chen2025janus} & AR &  0.89 & 0.79 &  0.66 & 0.80 & 86.90 & 89.32 & 84.19 \\
Bagel~\cite{deng2025emerging} & Diff. & 0.94 & 0.64 & 0.63 & 0.82 & - & - & - \\
Qwen-Image~\cite{wu2025qwen} & Diff. & 0.92 & 0.76 & 0.77 & 0.87 & 91.32 & 94.31 & 88.32 \\
\rowcolor[RGB]{234,242,255}
\textbf{\system} & AR & 0.91 & 0.75 & 0.60 & 0.79 & 89.85 & 92.00 & 84.48 \\
\rowcolor[RGB]{234,242,255}
\textbf{\system-RL} & AR & 0.93 & 0.89 & 0.90 & 0.86 & 90.14 & 92.08 &  86.00 \\
\bottomrule
\end{tabular*}
\end{table*}

In particular, \system obtains 87.3 on POPE~\cite{li2023evaluating} and 40.2 on MMMU~\cite{yue2024mmmu}, placing it on par with or above representative continuous unified models (\eg, Janus-Pro and Bagel) and clearly ahead of prior discrete unified models such as Emu3 and VILA-U. 
\system reaches 1463 on MME$_{\text{Perc}}$~\cite{fu2023mme} and 73.1 on SeedBench~\cite{li2024seed}, indicating strong capability on knowledge-intensive and reasoning-heavy queries. This suggests that our discrete visual representations preserve key semantics for recognition while remaining compatible with an autoregressive backbone that later supports generation and editing, avoiding the need for separate visual pathways. \system also achieves superior results on other benchmarks~\cite{liu2024mmbench, hudson2019gqa}.

\begin{table*}[!ht]
\centering
\small
\setlength{\tabcolsep}{0pt}
\renewcommand{\arraystretch}{1.1}
\caption{Comparison with state-of-the-art models on WISE benchmark for reasoning-based image generation.}
\vspace{-0.05in}
\label{tab:wise}
\begin{tabular*}{\linewidth}{@{\extracolsep{\fill}}l c c c c c c c c@{}}
\toprule
\textbf{Methods} & \textbf{Type} & \textbf{Cultural} & \textbf{Time} & \textbf{Space} & \textbf{Biology} & \textbf{Physics} & \textbf{Chemistry} & \textbf{Overall}$\uparrow$ \\
\midrule
Janus~\cite{wu2025janus} & AR    & 0.16 & 0.26 & 0.35 & 0.28 & 0.30 & 0.14 & 0.23 \\
VILA-U~\cite{wu2024vila} & AR    & 0.26 & 0.33 & 0.37 & 0.35 & 0.39 & 0.23 & 0.31 \\
SDv1.5~\cite{rombach2022high} & Diff. & 0.34 & 0.35 & 0.32 & 0.28 & 0.29 & 0.21 & 0.32 \\
Janus-Pro-7B~\cite{chen2025janus} & AR    & 0.30 & 0.37 & 0.49 & 0.36 & 0.42 & 0.26 & 0.35 \\
Emu3~\cite{wang2024emu3} & AR    & 0.34 & 0.45 & 0.48 & 0.41 & 0.45 & 0.27 & 0.39 \\
SD3 Medium~\cite{esser2024scaling}  & Diff. & 0.42 & 0.44 & 0.48 & 0.39 & 0.47 & 0.29 & 0.42 \\
SDXL~\cite{podell2023sdxl} & Diff. & 0.43 & 0.48 & 0.47 & 0.44 & 0.45 & 0.27 & 0.43 \\
PixArt-Alpha~\cite{chen2023pixart}         & Diff. & 0.45 & 0.50 & 0.48 & 0.49 & 0.56 & 0.34 & 0.47 \\
FLUX.1[Dev]~\cite{bfl_flux_2024}             & Diff. & 0.48 & 0.58 & 0.62 & 0.42 & 0.51 & 0.35 & 0.50 \\
BAGEL~\cite{deng2025emerging}      & Diff. & 0.44 & 0.55 & 0.68 & 0.44 & 0.60 & 0.39 & 0.52 \\
MetaQuery-XL~\cite{pan2025transfer}      & AR    & 0.56 & 0.55 & 0.62 & 0.49 & 0.63 & 0.41 & 0.55 \\
\rowcolor[RGB]{234,242,255}
\textbf{\system} & AR & 0.47 & 0.52 & 0.60 & 0.47 & 0.54 & 0.43 & 0.50 \\
\rowcolor[RGB]{234,242,255}
\textbf{\system-RL} & AR & 0.53 & 0.59 & 0.67 & 0.54 & 0.64 & 0.45 & 0.56 \\
\bottomrule
\end{tabular*}
\end{table*}

\begin{table*}[!ht]
\centering
\small
\setlength{\tabcolsep}{0pt}
\renewcommand{\arraystretch}{1.1}
\caption{Image editing results on GEdit-Bench. G\_SC (semantic consistency), G\_PQ (perceptual quality), and G\_O (overall score) refer to the metrics evaluated by GPT-4.1.}
\label{tab:gedit_bench}
\begin{tabular*}{\linewidth}{@{\extracolsep{\fill}}l c c c c c c@{}}
\toprule
\textbf{Model} &
\multicolumn{3}{c}{\textbf{GEdit-Bench-EN (Full set)}$\uparrow$} &
\multicolumn{3}{c}{\textbf{GEdit-Bench-CN (Full set)}$\uparrow$} \\
\cmidrule(lr){2-4}\cmidrule(lr){5-7}
& \textbf{G\_SC} & \textbf{G\_PQ} & \textbf{G\_O} & \textbf{G\_SC} & \textbf{G\_PQ} & \textbf{G\_O} \\
\midrule
AnyEdit~\cite{yu2025anyedit}                 & 3.18 & 5.82 & 3.21 & - & - & - \\
Instruct-Pix2Pix~\cite{brooks2023instructpix2pix} & 3.58 & 5.49 & 3.68 & - & - & - \\
MagicBrush~\cite{zhang2023magicbrush}           & 4.68 & 5.66 & 4.52 & - & - & - \\
UniWorld-v1~\cite{lin2025uniworld} & 4.93 & 7.43 & 4.85 & - & - & - \\
OmniGen~\cite{xiao2025omnigen}                 & 5.96 & 5.89 & 5.06 & - & - & - \\
BAGEL~\cite{deng2025emerging} & 7.36 & 6.83 & 6.52 & 7.34 & 6.85 & 6.50 \\
Step1X-Edit~\cite{liu2025step1x}          & 7.09 & 6.76 & 6.70 & 7.20 & 6.87 & 6.86 \\
\rowcolor[RGB]{234,242,255}
\textbf{\system} &  5.73 & 7.67 & 5.75 & 5.04 &  7.61 & 5.15 \\
\rowcolor[RGB]{234,242,255}
\textbf{\system-RL} & 6.85 & 7.68 & 6.68 & 6.38 & 7.67 & 6.27 \\
\bottomrule
\end{tabular*}
\end{table*}

\subsection{Image Generation Results}
We evaluate \system on three complementary benchmarks that probe different facets of text-to-image generation. As shown in Table~\ref{tab:geneval_dpg}, \system achieves competitive performance across GenEval~\cite{ghosh2023geneval} sub-dimensions, indicating reliable object-attribute binding and spatial control, and it attains high DPG~\cite{hu2024ella} scores on both global and relation metrics, suggesting that the generated images remain coherent while faithfully expressing interactions and relational constraints. WISE~\cite{niu2025wise} emphasizes reasoning-based generation, where correct outputs often require world knowledge across various domains. The results in Table~\ref{tab:wise} show that \system remains competitive among unified models and shows particularly strong results on time and space categories. This indicates improved ability to translate relational and structural constraints into visually correct outcomes. Importantly, this behavior is achieved within a single unified autoregressive model, in contrast to approaches that rely on specialized prompting or auxiliary mechanisms to strengthen reasoning-driven generation.

Reinforcement learning~\cite{guo2025deepseek} further improves generation results across these benchmarks. Notably, the discrete token interface makes preference optimization particularly straightforward, as it formulates multimodal generation as the same token-level optimization objective used for language models.

\subsection{Image Editing Results}
Table~\ref{tab:gedit_bench} reports image editing performance on GEdit-Bench~\cite{liu2025step1x}. In contrast to text-to-image benchmarks that evaluate generation from scratch, GEdit-Bench focuses on instruction-guided editing, where the model must apply the requested modification while keeping unrelated content unchanged and producing a visually coherent result. The benchmark summarizes these aspects with three GPT-based metrics~\cite{openai_gpt41_2025}: semantic consistency (G\_SC), perceptual quality (G\_PQ), and overall score (G\_O). Compared with prior approaches such as AnyEdit and UniWorld-v1, \system achieves substantially higher scores across metrics, indicating more reliable instruction execution and better preservation of source content.

Surprisingly, reinforcement learning yields a clear boost on image editing for \system, improving G\_O on the GEdit-Bench-EN full set from 5.75 to 6.68. By optimizing directly on preference feedback, it reduces common failure cases such as incomplete edits, excessive modifications, and attribute drift. This strong gain highlights the advantages of discrete token generation, where simple token-level preference optimization can lead to noticeably better visual outputs. 

\subsection{Analysis}
\noindent \textbf{Complementary supervision makes a unified visual tokenizer.} As discussed in Sec.~\ref{subsec:unified_tokenizer}, $\mathcal{L}_{cap}$ and $\mathcal{L}_{pix}$ provide supervision that is directly tied to downstream understanding and generation tasks, whereas $\mathcal{L}_{sig}$ and $\mathcal{L}_{feat}$ regularize the token space to preserve high-level semantics after discretization. To quantify their contributions, we take $\mathcal{L}_{\text{cap}}$ and $\mathcal{L}_{\text{pix}}$ as the base objectives, and ablate different loss combinations. ImageNet~\cite{deng2009imagenet} zero-shot accuracy, PSNR, codebook usage, and codebook perplexity are reported in Table~\ref{tab:tokenizer_loss_ablation}.

\begin{table}[!ht]
\centering
\small
\setlength{\tabcolsep}{0pt}
\renewcommand{\arraystretch}{1.1}
\caption{Ablation of tokenizer supervision objectives. We report ImageNet zero-shot accuracy (INet ZS), PSNR, codebook usage, and codebook perplexity (PPL).}
\label{tab:tokenizer_loss_ablation}
\vspace{-0.05in}
\begin{tabular*}{\linewidth}{@{\extracolsep{\fill}}c c c c c c c c@{}}
\toprule
$\mathbf{\mathcal{L}_{cap}}$ & $\mathbf{\mathcal{L}_{pix}}$ & $\mathbf{\mathcal{L}_{sig}}$ & $\mathbf{\mathcal{L}_{feat}}$ & \textbf{INet ZS}$\uparrow$ & \textbf{PSNR}$\uparrow$ & \textbf{Usage}$\uparrow$ & \textbf{PPL}$\uparrow$ \\
\midrule
\Checkmark   & \Checkmark  & \XSolidBrush & \XSolidBrush & 0.2 & 15.2 & 69.4 & 0.21 \\
\midrule
\Checkmark   & \Checkmark & \Checkmark & \XSolidBrush & 79.4 & 9.3	& 70.5 & 0.24\\
\Checkmark   & \Checkmark & \XSolidBrush & \Checkmark & 58.1 & 17.1	& 71.0 & 0.24 \\
\midrule
\Checkmark & \Checkmark & \Checkmark & \Checkmark & 80.2 & 19.6	& 75.6	& 0.28 \\
\bottomrule
\end{tabular*}
\end{table}

Training with only $\mathcal{L}_{cap}$ and $\mathcal{L}_{pix}$ leads to low ImageNet zero-shot accuracy. Although this metric is not a direct proxy for downstream VLM performance, the degraded codebook usage and perplexity suggest that the learned vocabulary is poorly utilized, resulting in a less expressive token space and weaker coverage of visual concepts. Adding $\mathcal{L}_{sig}$ or $\mathcal{L}_{feat}$ improves both recognition performance and codebook utilization, indicating that semantic regularization is crucial for building a compact yet expressive discrete vocabulary. Interestingly, $\mathcal{L}_{feat}$ improves reconstruction quality whereas $\mathcal{L}_{sig}$ slightly hurts PSNR, which may reflect a trade-off: enforcing stronger semantic clustering can discard some low-level appearance variations that are beneficial for pixel reconstruction. Finally, combining all objectives yields the best overall balance, achieving the highest ImageNet zero-shot score while also improving PSNR, usage, and perplexity.

\begin{figure}[!ht]
\centering
\begin{minipage}[t]{0.5\linewidth}
\centering
\includegraphics[width=\linewidth]{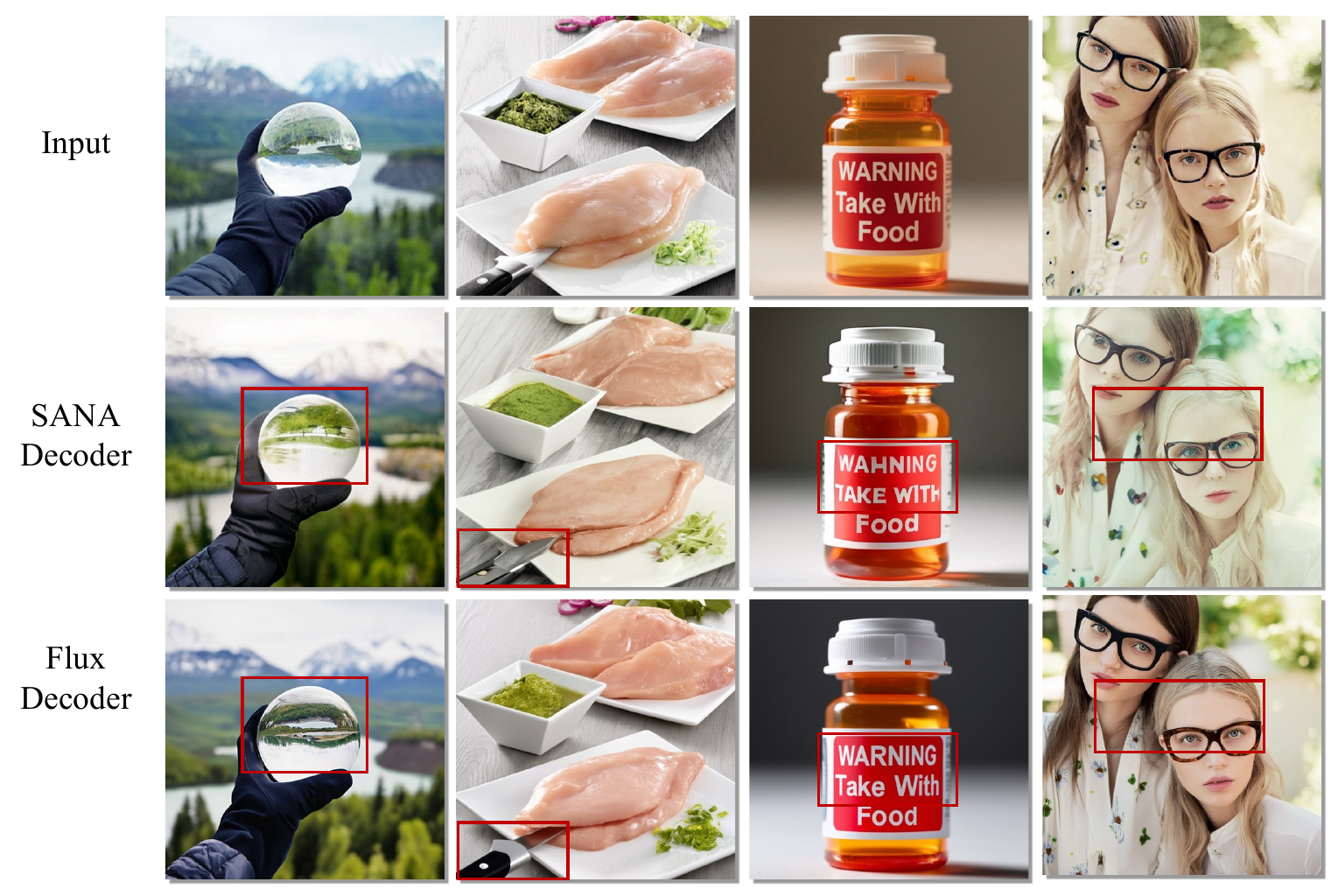}
\end{minipage}\hfill
\begin{minipage}[t]{0.5\linewidth}
\centering
\includegraphics[width=\linewidth]{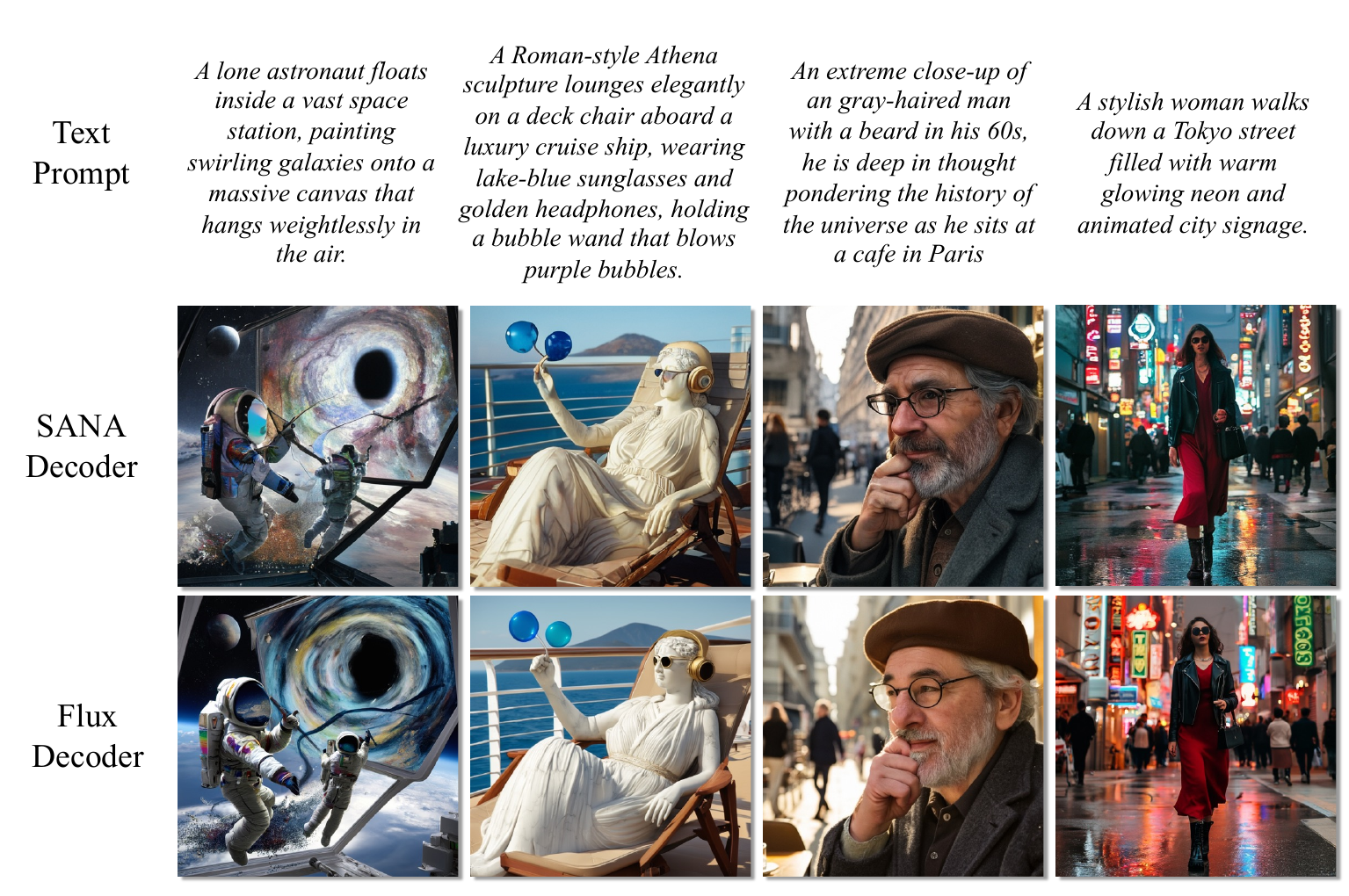}
\end{minipage}
\caption{Comparison between diffusion decoders. \textbf{Left:} reconstruction comparison between SANA1.5 and Flux with shared visual tokens. \textbf{Right:} text-to-image comparison between diffusion decoders.}
\label{fig:decoder_comparison}
\end{figure}

\vspace{0.05in}
\noindent \textbf{LMM generates, diffusion model renders.} We compare our chosen diffusion decoder, FLUX.1[Dev]~\cite{bfl_flux_2024}, to a smaller decoder, Sana1.5-1.6B~\cite{xie2025sana}, while keeping the visual tokenizer fixed. As shown in Figure~\ref{fig:decoder_comparison}, both decoders reconstruct highly similar images, suggesting that the visual tokens encode not only the global layout and object composition, but also much of the low-level details.  Meanwhile, the diffusion decoder primarily serves as a renderer that maps the predicted visual tokens back to pixel space. 

Interestingly, we observe that a stronger decoder is more robust on challenging patterns such as faces and text, where visual quality is particularly sensitive to subtle artifacts. The gap between decoders becomes even more pronounced for text-to-image generation: the FLUX decoder produces sharper details and more stable typography, while the SANA decoder exhibits more frequent texture blur and character distortions. This suggests the decoder capacity mainly dominates visual fidelity, while the high-level semantics are largely determined by the discrete tokens predicted by autoregressive models.

\noindent \textbf{Semantic tokenizer reduces the reliance on classifier-free guidance.}
Prior autoregressive image generation models~\cite{sun2024autoregressive} built on VQ-VAE~\cite{esser2021taming} often depend heavily on classifier-free guidance (CFG)~\cite{ho2022classifier} to obtain prompt-faithful outputs, whereas language models typically require no analogous mechanism. We hypothesize this gap largely stems from the semantics of the visual vocabulary: VQ-VAE tokens are optimized primarily for reconstruction and not aligned to the language modality, making the conditional signal less explicit at sampling time. 

\begin{figure}[!ht]
\centering
\includegraphics[width=0.7\linewidth]{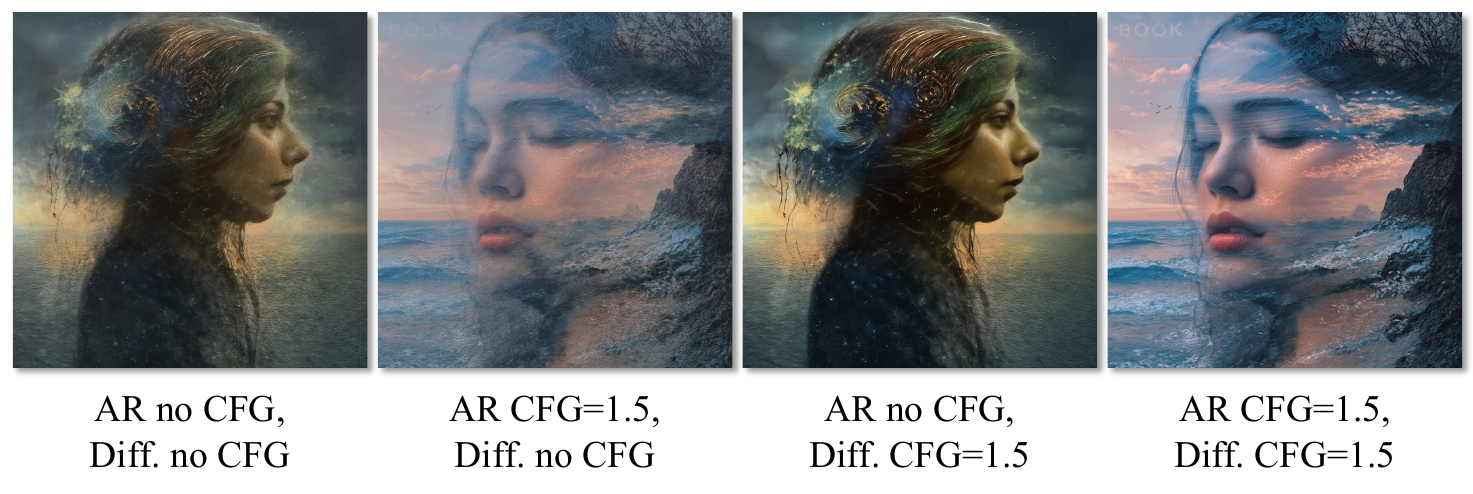}
\caption{Image generation w/ and w/o CFG. Prompt: ```Book cover, A surreal double exposure portrait that blends a woman’s face with a beautiful seascape'''.}
\label{fig:cfg}
\end{figure}

To verify this, we explore the effects of disabling CFG in \system during inference. The comparison in Figure~\ref{fig:cfg} shows that our model still produces images that follow the instruction and remain visually coherent. Enabling CFG yields marginal gains, mainly improving overall smoothness and suppressing minor artifacts. We also disable CFG in the diffusion decoder, and the results show that generation quality only degrades in local textures (\eg, hair). This observation supports our previous claim that the predicted visual tokens already provide a strong conditional signal, while the diffusion decoder mainly contributes pixel rendering. Overall, autoregressive generation with a semantic tokenizer makes it possible to use weaker guidance or even remove CFG altogether, which can significantly accelerate inference by avoiding extra forward passes.

\begin{table*}[t] 
\centering 
\small 
\setlength{\tabcolsep}{0pt} 
\renewcommand{\arraystretch}{1.1} \caption{Comparison on different RL recipes. Generative RL induces T2I–Editing synergy while preserving multimodal understanding across benchmarks. T2I RL, Edit RL, and Joint RL refer to reinforcement learning over text-to-image examples, image editing examples, and a mixture of both. This table reports the G\_O score for GEdit-Bench-EN, GenEval and WISE overall scores, and the DPG overall score.} 
\begin{tabular*}{\linewidth}{@{\extracolsep{\fill}}lccccccccccc@{}} 
\toprule \multirow{2}{*}{\textbf{RL Recipes}}  & \multicolumn{1}{c}{\textbf{Editing}}  & \multicolumn{3}{c}{\textbf{T2I}}  & \multicolumn{7}{c}{\textbf{Understanding}} \\  \cmidrule(lr){2-2} \cmidrule(lr){3-5} \cmidrule(lr){6-12}   & \textbf{GEdit}  & \textbf{GenEval}  & \textbf{DPG}  & \textbf{WISE}  & \textbf{POPE}  & \textbf{MMB}  & \textbf{MME}  & \textbf{MMMU}  & \textbf{GQA}  & \textbf{VQAv2}  & \textbf{SEED} \\ 
\midrule \textbf{\system}-SFT & 5.75 & 0.79 & 84.48 & 0.50  & 87.28 & 80.67 & 1462.89 & 40.2 & 59.83 & 76.11 & 73.07 \\  
$\rightarrow$ T2I RL & 5.92 & 0.85 & 85.57 & 0.55  & 87.56 & \textbf{80.99} & 1449.17 & \textbf{41.4} & 60.25 & 76.10 & 73.16 \\  
$\rightarrow$ Edit RL & 6.31 & 0.80 & 84.53 & 0.54  & 87.09 & 80.74 & 1454.72 & 40.1 & \textbf{60.28} & \textbf{76.17} & \textbf{73.41} \\  
$\rightarrow$ T2I RL $\rightarrow$ Edit RL & 6.42 & 0.84 & \textbf{86.03} & \textbf{0.56}  & \textbf{87.65} & 80.83 & 1451.22 & 41.0 & 60.14 & 75.93 & 73.33 \\  
$\rightarrow$ T2I RL $\rightarrow$ Edit RL $\rightarrow$ Joint RL & \textbf{6.68} & \textbf{0.86} & 86.00 & \textbf{0.56}  & 87.11 & 80.69 & \textbf{1464.48} & 41.0 & 60.17 & 75.91 & 72.97 \\  
\bottomrule 
\end{tabular*}  
\label{rl_ablation} 
\end{table*}

\begin{figure*}[!ht]
\centering
\includegraphics[width=\linewidth]{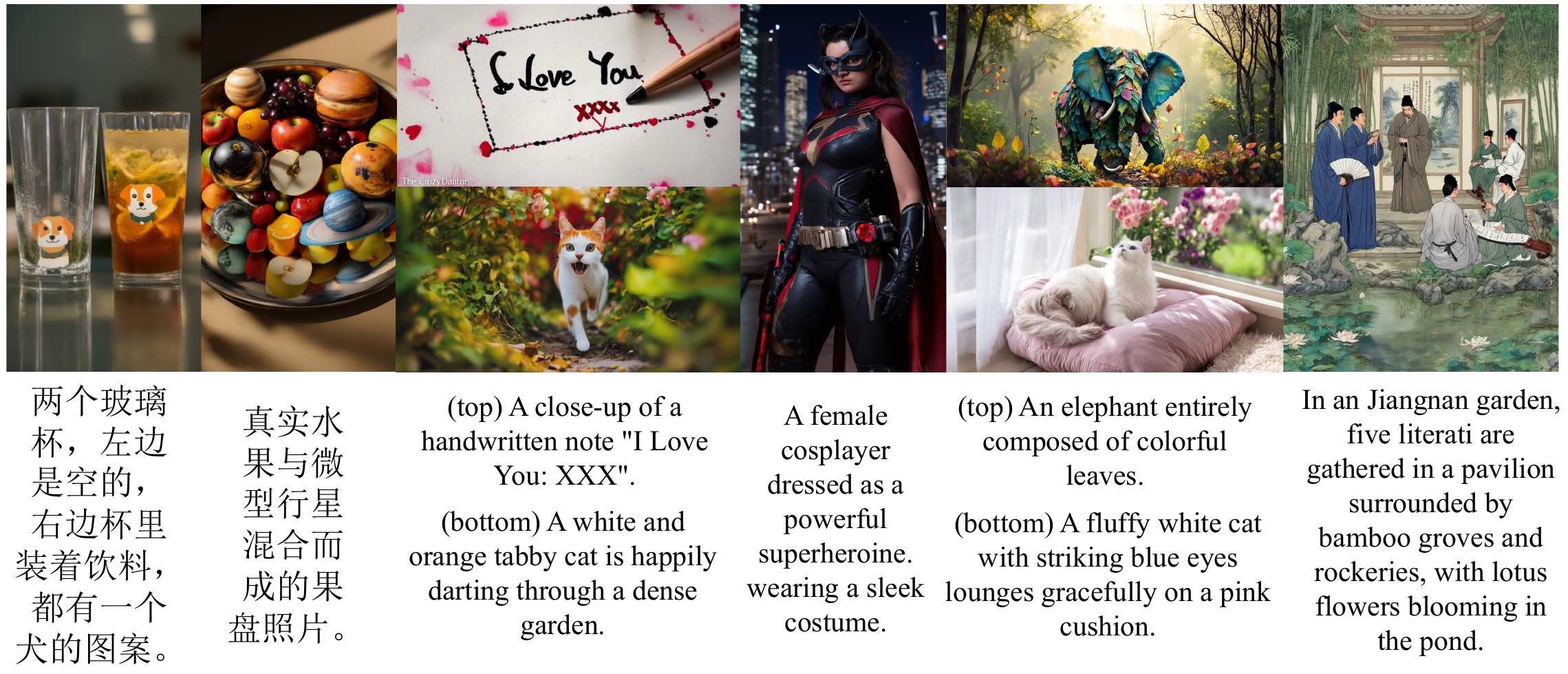}
\includegraphics[width=\linewidth]{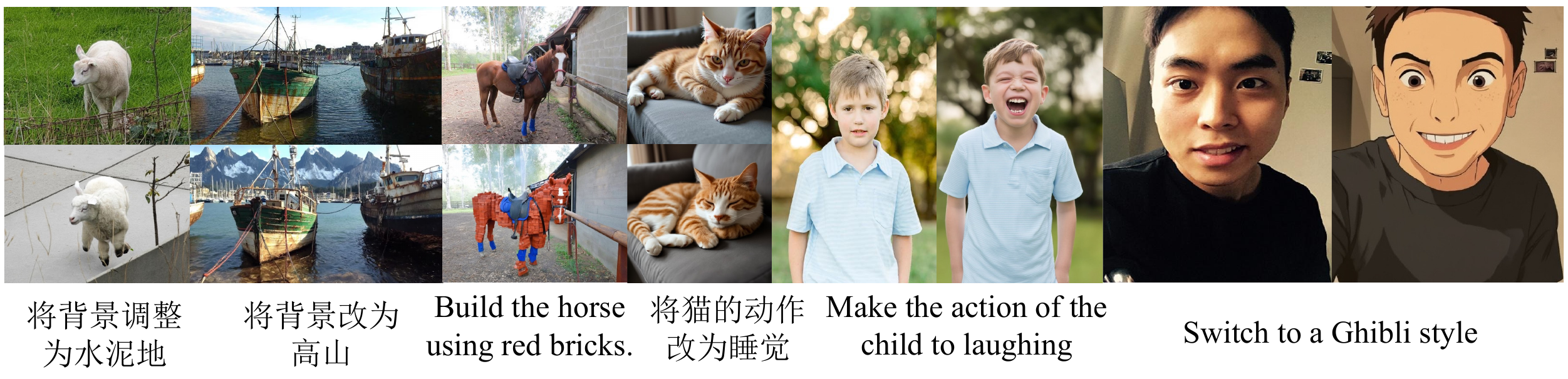}
\caption{Image generation and editing results by \system.}
\label{fig:visualization}
\end{figure*}

\noindent \textbf{RL on \system demonstrates cross-task synergy.} This section discusses several key findings during the reinforcement learning (RL) stage, with numbers reported in Table~\ref{rl_ablation}. First, we conduct RL training independently for text-to-image (T2I) and image editing tasks. The results show performance gains for the target task and, remarkably, consistent improvements in the reciprocal task. For instance, T2I RL improves the GEdit score from 5.75 to 5.92, while after Edit RL, the GenEval score increases from 0.79 to 0.80. Crucially, we find that multimodal understanding performance remains stable across all benchmarks during these individual RL stages. Based on these observations, we initialize the model with the T2I RL weights for subsequent Edit RL, followed by a final stage of joint RL. Experimental results demonstrate that performance on all tasks can be further improved, with the final joint RL stage achieving the highest overall scores. 

The above experiments highlight a fundamental advantage of the unified architecture adopted by \system: by modeling disparate tasks through a shared visual latent space, policy updates facilitate constructive accumulation rather than mutual interference, where gains on one task naturally propagate to others without incurring a performance penalty.

\noindent \textbf{Visualizations.} We provide visualization results by \system in Figure~\ref{fig:visualization}. \system can maintain natural and structured layouts in complex scenes, including both narrative object compositions and fine-grained environment details. For instruction-guided editing, the examples below show that \system can follow compound edit requests that mix appearance edits, pose changes, and style transfer in a single instruction, indicating strong controllability under interleaved text and visual tokens.

\section{Conclusion}
This paper presents \system, an autoregressive large multimodal model built on unified discrete visual representations. We train a semantic tokenizer that discretizes images into compact token sequences that preserve both language-aligned semantics and visual details. With this tokenizer, we scale autoregressive training on large-scale multimodal tokens using a 7B model. Finally, we apply Group Relative Policy Optimization to further align model outputs with preference feedback for generation and editing. Extensive experimental results demonstrate the potential of next-token prediction in unifying various multimodal tasks with competitive performance.

\bibliographystyle{plainnat}
\bibliography{main}

\end{document}